\documentclass[journal]{IEEEtran}

\usepackage{graphicx}
\usepackage{overpic}
\usepackage{amsmath}
\usepackage{flushend}
\usepackage{amssymb}
\usepackage{multirow}
\usepackage{makecell}
\usepackage{array}
\usepackage{stfloats}
\usepackage{hyperref}
\begin{document}
\title{SAR Image Change Detection Based on Multiscale Capsule Network}
\author{Yunhao Gao, Feng Gao, Junyu Dong, Heng-Chao Li
\thanks{This work was partially supported by the National Natural Science Foundation of China (No. 41606198, 41576011).}
\thanks{Y. Gao, F. Gao, and J. Dong are with the Qingdao Key Laboratory of Mixed Reality and Virtual Ocean, School of Information Science and Engineering, Ocean University of China, Qingdao 266100, China.

H.-C. Li is with the Sichuan Provincial Key Laboratory of Information Coding and Transmission, Southwest Jiaotong University, Chengdu 610031, China.}
}

\markboth{Submitted to IEEE Geoscience and Remote Sensing Letters}%
{Shell}

\maketitle
\begin{abstract}
Traditional change detection methods based on convolutional neural networks (CNNs) face the challenges of speckle noise and deformation sensitivity for synthetic aperture radar images. To mitigate these issues, we proposed a \underline{M}ulti\underline{s}cale \underline{Caps}ule \underline{Net}work (Ms-CapsNet) to extract the discriminative information between the changed and unchanged pixels. On the one hand, the capsule module is employed to exploit the spatial relationship of features. Therefore, equivariant properties can be achieved by aggregating the features from different positions. On the other hand, an adaptive fusion convolution (AFC) module is designed for the proposed Ms-CapsNet. Higher semantic features can be captured for the primary capsules. Feature extracted by the AFC module significantly improves the robustness to speckle noise. The effectiveness of the proposed Ms-CapsNet is verified on three real SAR datasets. The comparison experiments with four state-of-the-art methods demonstrated the efficiency of the proposed method. Our codes are available at \href{https://github.com/summitgao/SAR\_CD\_MS\_CapsNet}{\it{https://github.com/summitgao/SAR\_CD\_MS\_CapsNet}}.

\end{abstract}

\begin{IEEEkeywords}
Change detection, multiscale capsule network, synthetic aperture radar, deep learning.
\end{IEEEkeywords}

\IEEEpeerreviewmaketitle

\section{Introduction}

\IEEEPARstart{S}{ynthetic} aperture radar (SAR) imaging acquisition technologies have been developed rapidly. Plenty of multitemporal SAR images are available to monitor the changed information of the earth. Therefore, SAR image change detection has drawn increasing attention recently. Researchers have designed a variety of SAR change detection methods for ecological surveillance, disaster monitoring \cite{Brunner10_tgrs}, and urban planning \cite{Quan18_jstars}. 

Although plenty of techniques have been proposed \cite{Radke05_tip}, SAR image change detection is a still challenging task. Image quality is deteriorated by speckle noise which hinders the meticulous interpretation of SAR data. Many methods are implemented to address the issue of speckle noise. They are usually comprised of three steps: 1) image coregistration, 2) difference image (DI) generation, and 3) DI classification \cite{Gong16_tnnls}. Image coregistration is a fundamental task for SAR image change detection. The spatial correspondences between multitemporal SAR images can be established. In the second step, the DI is commonly generated by the log-ratio, Gauss-ratio \cite{Hou14_jstars}, and neighborhood-ratio \cite{Gong12_grsl} operators. For the DI classification step, most researches are devoted to building a robust classifier. It is a non-trivial task since a powerful classifier directly determines the precision of change detection. 

Many researchers are dedicated to developing powerful classifiers for change detection. Li {\it et al.} \cite{Li15_grsl} designed two-level clustering algorithm for unsupervised change detection. In \cite{Jia16_grsl}, local-neighborhood information is embedded in the clustering objective function to improve the change detection performance. Gong {\it et al.} \cite{Gong12_tip} developed an improved Markov random field (MRF) based on fuzzy $c$-means (FCM) clustering to suppress the speckle noise. In \cite{Gong16_tnnls}, a deep belief network (DBN) was employed for SAR image change detection. Some of the most noteworthy approaches also achieved breakthrough relying on the level-set algorithm \cite{Bazi10_tgrs}, stacked autoencoder (SAE) \cite{Planinsic18_grsl}, PCANet \cite{Gao16_grsl}. 

In recent years, the convolutional neural network (CNN) has greatly boosted the performance of many visual tasks \cite{Girshick14_cvpr}. It is widely acknowledged that CNN is capable of robust feature learning. Inspired by these achievements, CNN has been successfully applied in SAR change detection \cite{Liu18_tgrs}. Zhan {\it et al.} \cite{Zhan17_grsl} refined a deep siamese CNN to measure the similarity of patch-pair from SAR images. In \cite{Gao19_grsl}, transferred deep learning was applied to sea ice SAR image change detection based on CNN. Liu {\it et al.} \cite{Liu19_tnnls} proposed an elegant local restricted CNN (LR-CNN) for polarimetric SAR change detection. Although CNN-based methods have achieved excellent performance in SAR image change detection, the accuracy sometimes deteriorates under the case of transformation, such as tilts and rotations. Specifically,  CNN is incapable of modeling the positional relationship among ground objects.

More recently, the capsule network (CapsNet) was designed for assigning parts to wholes \cite{Hinton17_nips}. An activity vector from capsules represents the entity instantiation parameters such as pose, texture, and deformation. The existence of entities is expressed by the length of instantiation parameters. To propagate information, dynamic routing mechanism is utilized to send activity vectors to the appropriate layer above. It is empirically verified that the CapsNet is effective for image segmentation and classification tasks. A few attempts based on the CapsNet have been applied for remote sensing images analysis, and obtain promising results. In \cite{Paoletti19_tgrs}, a new CNN architecture based on capsule networks is proposed for hyperspectral image (HSI) classification. Zhu {\it et al.} \cite{Zhu19_remotesens} proposed a 3D deep capsule network for HSI classification. The local connection and weight sharing strategy greatly reduces the number of parameters. As far as we know, the literature on the CapsNet-based SAR change detection is
very sparse. 

We argue that the weakness of existing SAR image change detection approaches mainly comes from two aspects: One is the correlation of features from different positions fails to be modeled effectively. The other one lies in the intrinsic speckle noise in SAR images. With the aforementioned challenges, we proposed a \underline{M}ulti\underline{s}cale \underline{Caps}ule \underline{Net}work (Ms-CapsNet) to extract the discriminative information between multitemporal SAR images. To enhance the feature correlations, we introduced the capsule module to achieve the equivariant properties. The proposed Ms-CapsNet provides a group of instantiation parameters which can effectively capture features from different positions. To tackle the problem of speckle noise, an adaptive fusion convolution (AFC) module is designed to convert pixel intensities to activities of the local features. Accordingly, local features are noise robustness. Extensive experiments on three real datasets are conducted to show the superiority of our proposed method over four state-of-the-art works.

For clarity, the main contributions are summarized as follows:

\begin{itemize}
\item The multiscale capsule network (Ms-CapsNet) is designed to extract the features from different positions. And equivariant properties can be achieved by capsule module. Logically, the demand for training samples is reduced by the correlative and complete information to some extent.

\item A simple yet effective AFC module is designed, which can effectively convert pixel intensities to activities of local features. The AFC module extracts higher semantic features and emphasizes the meaningful one through attention-based strategy. Therefore, the activity local features become more noise robustness and immediately accepted as the input of the primary capsule.

\item Extensive experiments have been implemented on three SAR datasets to validate the effectiveness of the proposed method. Moreover, we have released the codes and setting to facilitate future researches in multitemporal remote sensing image analysis.

\end{itemize}

\section{Methodology}

\begin{figure*}
\centering
\begin{center}
\includegraphics [width=6in]{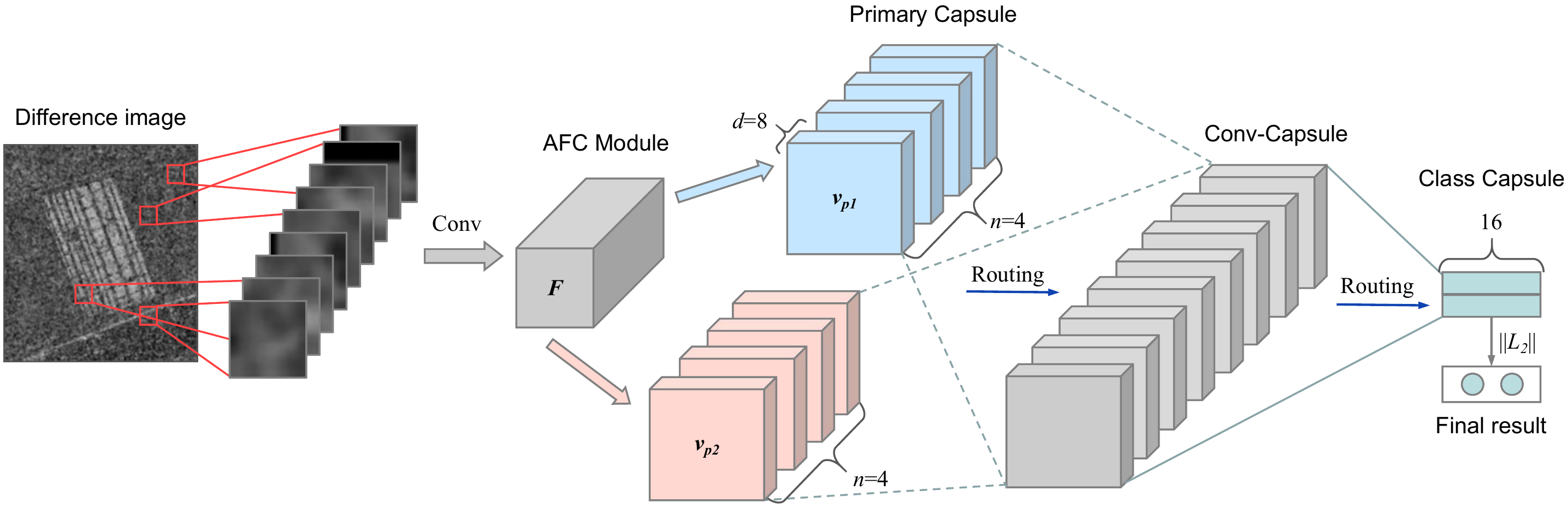}
\caption{Illustration of the proposed change detection method based on multiscale capsule network (Ms-CapsNet). First, image patches are fed into AFC module for higher semantic features. Then, multiscale primary capsules layers are adopted with kernel $3\times 3$ and $5\times 5$ to obtain primary capsules $v_{p1}$ and $v_{p2}$. Later $v_{p1}$ and $v_{p2}$ are input to conv-capsule layer and class capsule layer, respectively. Finally, fuse the output of class capsule layers to calculate the final result.}
\label{fig_capsnet}
\end{center}
\end{figure*}

The proposed method is illustrated in Fig. \ref{fig_capsnet}. A difference image (DI) is first generated by the log-ratio operator. Hierarchical FCM clustering \cite{Gao16_grsl} is employed to select reliable training samples for Ms-CapsNet. Finally, pixels in the DI are classified by the trained Ms-CapsNet to obtain the final change map.

In our implementations, image patches are extracted from DI with the size of $r \times r$. The proposed Ms-CapsNet is comprised of AFC and capsule modules. The AFC module is used to convert pixel intensities to high semantic features through which the speckle noise is suppressed to some extent. The capsule module is utilized to activate high semantic features. In the following subsections, we will describe both modules in detail. 

\subsection{Adaptive Fusion Convolution Module}

\begin{figure}[t]
\begin{center}
\includegraphics [width=2.5in]{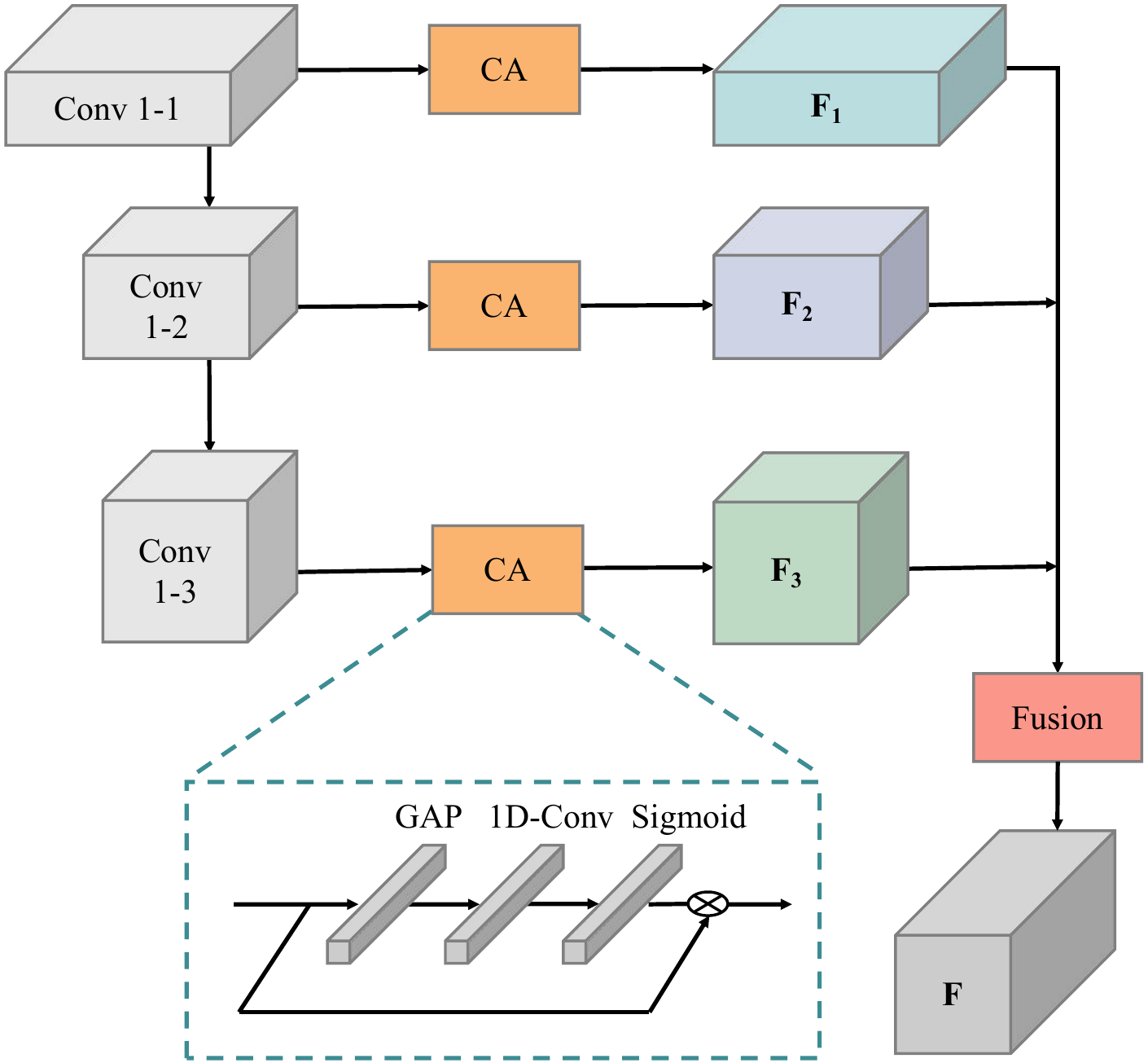}
\caption{Illustration of the Adaptive Fusion Convolution (AFC) module.}
\label{fig_afc}
\end{center}
\end{figure}

As shown in Fig. \ref{fig_afc}, the proposed AFC module is utilized to encode the input. Recently, the self-attention mechanism has been employed for visual structure understanding. The long-range dependencies can be captured by the self-attention mechanism. Inspired by Hu's work \cite{Hu18_cvpr}, we design a simple yet effective AFC module based on channel-wise attention (CA). First, atrous convolution (Conv 1-1, Conv 1-2 and Conv 1-3 with kernel size $3\times 3$) is adopted with different dilation rates, which are set to $1$, $2$ and $3$ to capture multiscale features. Then, the multiscale features are aggregated by CA-based feature fusion.

The input features $\textbf{F}_{in}\in\mathbb{R}^{w_0\times w_0 \times c_0}$ from atrous convolution are fed into CA. Then, global average pooling (GAP) squeezes $\textbf{F}_{in}$ in the spatial domain to obtain $\textbf{F}_{avg} \in \mathbb{R}^{1\times 1\times c_0}$. Then, 1D-Convolution (1D-Conv) is employed to explore the channel relationship of $\textbf{F}_{avg}$. After the $Sigmoid$ function, a channel weighting-based vector $M$ can be obtained. Finally, the channel weighting-based feature $\textbf{F}_{out}$ can be computed as $\textbf{F}_{out}=M\otimes F_{in}$, where $\otimes$ denotes channel-wise multiplication. Therefore, the channel weighting-based features from Conv 1-1, Conv 1-2 and Conv 1-3 are $\textbf{F}_{1}$, $\textbf{F}_{2}$ and $\textbf{F}_{3}$. We fused the features by pixel-wise summation:

\begin{equation}
\textbf{F} = D_1(\textbf{F}_1)+ D_2(\textbf{F}_2)+D_3(\textbf{F}_3),
\end{equation}
where $\textbf{F}$ represents the fused features, $D_1$, $D_2$, and $D_3$ are dimension matching functions which are operated by $1\times 1$ convolution.

\subsection{Capsule Module}

The capsule module is a neural network comprised of the primary capsule layer, the conv-capsule layer, and the fully-connected layer, as illustrated in Fig. \ref{fig_capsnet}.

\subsubsection{Primary Capsule Layer}

This layer is employed to extract the low-level features from multi-dimensional entities through convolutional-like operation with kernel size $k\times k$. Different from traditional convolution, multiple feature maps will be obtained instead of one. The primary capsule layer first receive the feature map $F\in \mathbb{R}^{w\times w \times c}$ from the AFC module. Then convolutional-like operation and $squashing$ activation function are employed to obtain the output capsules $v_{p}$. The $squashing$ activity function is denoted as:

\begin{equation}
\label{squash}
v=\frac{\|s\|^2}{1+\|s\|^2}\frac{s}{\|s\|},
\end{equation}
where $s$ is the total input and $v$ is the vector output of capsule. In the primary capsule layer, the size of the output capsules $v_p$ is $w_1\times w_1\times n \times d$, where $n$ is the number of feature maps, $n\times d =c$ and $d=8$. The $[w_1\times w_1]$ grid is shared weights. 
In other words, we obtain $[w_1\times w_1 \times n]$ 8D vectors in total primary capsules. In our implementations, multi-scale information is taken into account. Two primary capsule layers are employed with kernel size $k=3$ and $k=5$, respectively. Therefore, multi-scale feature representation can be obtained. Feature vectors from two scales are denoted by $v_{p1}$ and $v_p$, respectively. 

\subsubsection{Conv-Capsule Layer} 

This layer uses local connections and the shared transformation matrix to reduce the number of parameters to some extent \cite{Zhu19_remotesens}. Conv-capsule layer uses the dynamic routing strategy to update the coupling coefficient $c$. The connection (transformation matrix) between the primary capsule layer and the conv-capsule layer is $W$, and the transformation matrix $W$ is also shared in each grid. Therefore, the output $v_c$ of the conv-capsule layer can be expressed as:

\begin{equation}
v_c=squashing(\sum c\cdot u), 
\end{equation}
where $c$ is the coupling coefficient, $u=W\cdot v_{p}$. $v_{p}$ is the output of the primary capsule layer. For dynamic routing, we first set the agreement $b$ to $0$. The coupling coefficient $c$ can be calculated by $c = softmax(b)$. That is to say, we update $b$ to calculate the latest coupling coefficient $c$. In addition, the update process of $b$ can be expressed as $b\gets b+u \cdot v_c$. The detailed descriptions of the dynamic routing can be found in \cite{Hinton17_nips}.

\subsubsection{Class Capsule Layer} 

The class capsule layer can be considered as a fully connected layer. Dynamic routing mechanism is still used for coupling coefficient updating. In this layer, multiscale activity vectors $v_{o1} \in \mathbb{R}^{2\times 16}$ and $v_{o2} \in \mathbb{R}^{2\times 16}$ from class capsule layer are fused by summation $v_o=v_{o1}\oplus v_{o2}$. Then the vector norm is calculated to measure the probability of classes. The loss function of Ms-CapsNet can be defined as:

\begin{equation}
\begin{split}
\label{loss}
L=T_k \max(0, m^+- &\|v_o\|)^2+ \\
& \lambda(1-T_k)\max (0, \|v_o\|-m^-)^2.
\end{split}
\end{equation}
Here $T_k = 1$ when the label $k$ is presented ($k=0$ means the unchanged class, $k = 1$ means the changed class). $\lambda=0.5$ is used to constrain the length of the activity vector of the initial class capsule. If there is a changed class object in the image, the class capsule of the changed class should output a vector with a length of at least $m^+=0.9$. On the contrary, if there is no object of the changed class, a vector with a length less than $m^-=0.1$ will be output from the class capsule. Then, the final change map can be calculated by pixel-wise classification. 

\section{Experimental Results and Analysis}

In this section, we first describe the datasets and evaluation criteria in our experiments. Next, an exhaustive investigation of several vital parameters on the change detection performance is presented. Finally, we conduct extensive experiments to verify the effectiveness of the proposed method.

\begin{figure*}[htbp]
\centering
\includegraphics [width=7in]{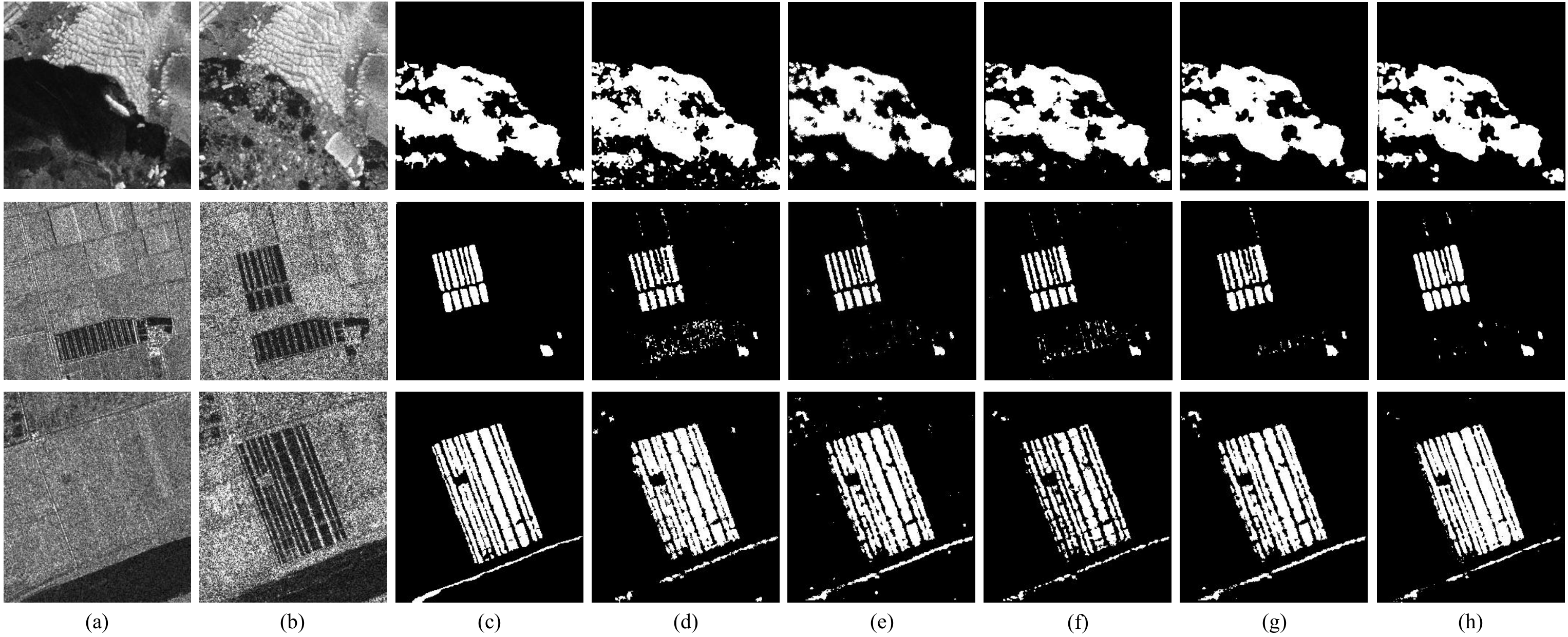}
\caption{Visualized results of different change detection methods on three datasets. (a) Image captured at $t_1$. (b) Image captured at $t_2$. (c) Ground truth change map. (d) Result by PCANet. (e) Result by MLFN. (f) Result by DCNN. (g) Result by LR-CNN. (h) Result by the proposed Ms-CapsNet.}
\label{fig_result}
\end{figure*}

\subsection{Dataset and Evaluation Criteria}

To verify the effectiveness of the proposed Ms-CapsNet, we employed Ms-CapsNet on three multitemporal SAR datasets acquired by different sensors. The first dataset is the Sulzberger dataset. It is captured at Sulzberger Ice Shelf by Envisat satellite of the European Space Agency on March 11 and 16, 2011, respectively. The size of the dataset is $256\times256$ pixels, as illustrated in the first row of Fig. \ref{fig_result} (a)-(c). The other two datasets named Yellow River I and Yellow River II datasets, are captured at the Yellow River Estuary by Radarsat-2 in June 2008 and June 2009, respectively. Their sizes are $257\times289$ and $306\times291$ pixels in the second and third rows of Fig. \ref{fig_result} (a)-(c), respectively. It is very challenging to perform accuracy change detection on the Yellow River dataset since the speckle noise is much stronger. Geometric corrections have been performed on these datasets, and the ground truth images were manually annotated carefully with expert knowledge.

In the following experiments, the proposed Ms-CapsNet is compared with four closely related methods, including the PCANet \cite{Gao16_grsl}, the transferred multilevel fusion network (MLFN) \cite{Gao19_grsl}, the deep convolutional neural networks (DCNN) \cite{Song18_tgrs}, the CNN with local spatial restrictions (LR-CNN) \cite{Liu19_tnnls}. To verify the effectiveness of the proposed Ms-CapsNet, false positives (FP), false negatives (FN), percentage correct classification (PCC), overall errors (OE), and Kappa coefficient (KC) are adopted as the evaluation criteria.

\subsection{Parameters Analysis of the Proposed Ms-CapsNet}

\subsubsection{Analysis of the Patch Size}

\begin{figure}[ht]
\centering
\includegraphics [width=3in]{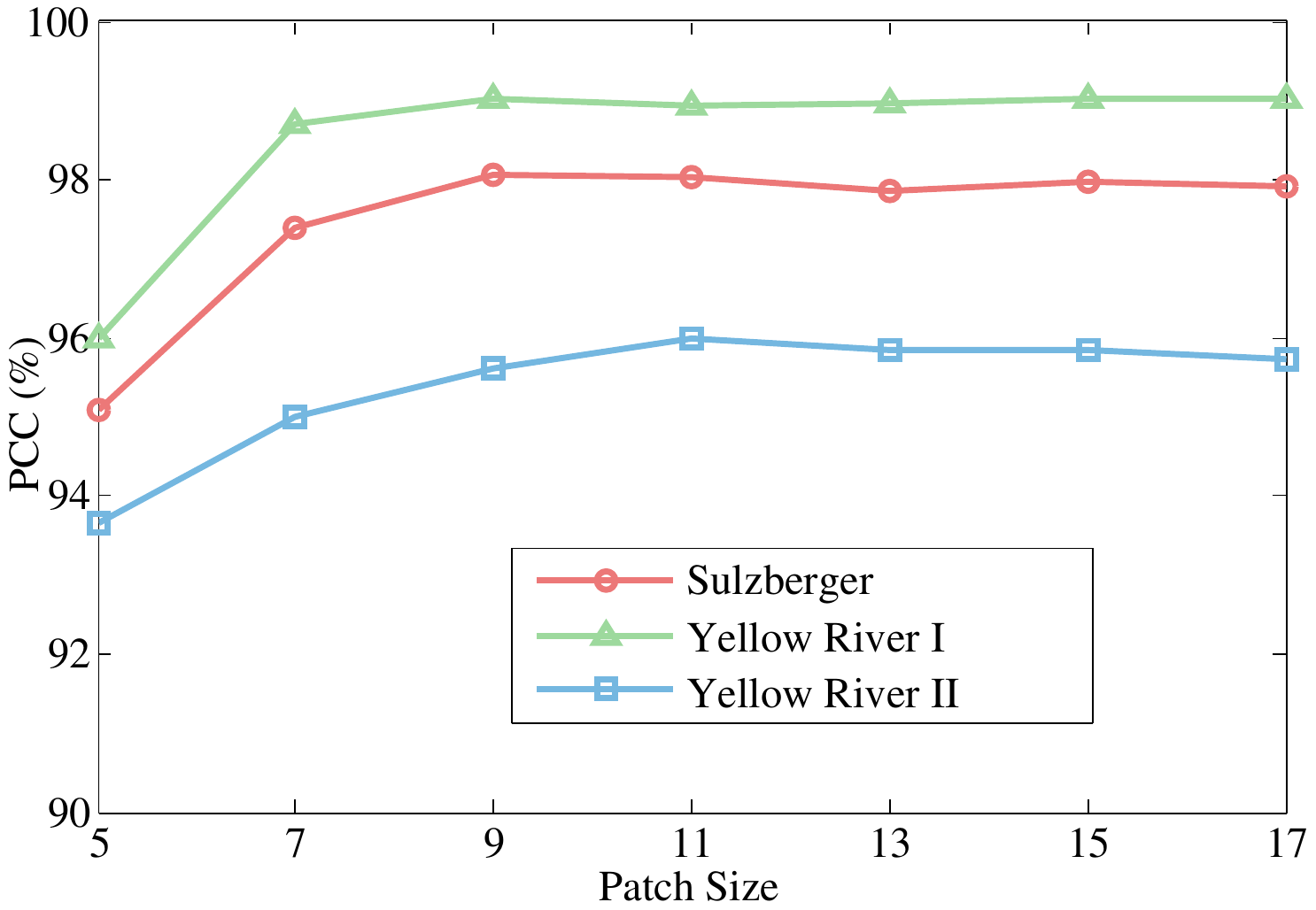}
\caption{Relationship between the PCC value and patch size.}
\label{fig_para}
\end{figure}

The patch size $r$ represents the scale of spatial neighborhood information. Fig. \ref{fig_para} shows the relationship between $r$ and PCC, where $r$ is changed from 5 to 17. According to Fig. \ref{fig_para}, the PCC values increase first and then tend to be stable. It is evident that the contextual information is important for change detection. However, a large patch size will increase the computational cost. Therefore, we choose $r=9$ for the Sulzberger and Yellow River I datasets, and $r=11$ for the Yellow River II dataset.

\subsubsection{Analysis of The Training Sample Numbers}

\renewcommand\arraystretch{1.2}
\begin{table}[h]
\centering 
\caption{Relationship between the PCC value and the number of training samples.}
\begin{tabular}{ l |p{0.8cm}<{\centering} p{0.8cm}<{\centering} p{0.8cm}<{\centering} p{0.8cm}<{\centering} p{0.8cm}<{\centering}}
\hline \hline

\multirow{2}{*}{Method} &

\multicolumn{5}{c}{PCC of different training samples number (\%)}\\
\cline{2-6} 
 & 200 & 400 & 600 & 800 & 1000 \\ \hline  
 
 PCANet \cite{Gao16_grsl}  & 91.44   & 92.12  & 93.34  & 93.88 & 94.26 \\   
 MLFN \cite{Gao19_grsl}    & 94.20    &94.72   & 95.01  & 95.24 & 95.48\\ 
 DCNN \cite{Song18_tgrs} & 92.51    & 92.58   & 92.66  & 93.07  & 93.81 \\ 
 LR-CNN \cite{Liu19_tnnls} &93.81    &94.24   &94.68  &95.00 &95.32 \\ 
 Ms-CapsNet     & 94.54    & 94.98  & 95.28  & 95.45 & 95.78 \\

 \hline \hline
\end{tabular}
\label{table_para}
\end{table}

Table \ref{table_para} compares the proposed Ms-Capsule with other methods when considering different number of training samples, i.e, $200$, $400$, $600$, $800$, and $1000$. We can observe that the accuracy of other methods drops sharply when the number of samples is less than $600$. Especially, DCNN and LR-CNN depend heavily on large volumes of training data, and few training samples will lead to overfitting which degrades the performance. In summary, the PCC value of the proposed method is less afflicted with the training sample numbers. It is because the feature spatial correlations can reduce the dependence on training samples to some extent.

\subsection{Change Detection Results on Three Datasets}

\renewcommand\arraystretch{1.2}
\begin{table*}[htbp]
\centering 
\caption{Change detection results on three datasets.}
\begin{tabular}{ l |c c c c c | c c c c c | c c c c c}
\hline \hline

\renewcommand\arraystretch{2}
\multirow{2}{*}{Method} &
\multicolumn{5}{c|}{Sulzberger dataset}&
\multicolumn{5}{c|}{Yellow River I dataset}&
\multicolumn{5}{c} {Yellow River II dataset}\\
\cline{2-16} 
 & FP & FN & OE & PCC & KC & FP & FN & OE & PCC & KC & FP & FN & OE & PCC & KC \\ \hline  
 
 PCANet \cite{Gao16_grsl}   & 1410   & 1437  & 2847 & 95.66 & 88.63  &
 1063  & 362  & 1425  & 98.40 & 86.47 & 
 2435   & 1533 & 3968  & 94.66 &82.43 \\   
 MLFN \cite{Gao19_grsl}    & 616    & 664   & 1280  & 98.05 & 94.89 & 
 721    & 863   & 1584  & 98.22 & 83.82 & 
 1544   & 1972  & 3516  & 95.27 & 83.82\\ 
 DCNN \cite{Song18_tgrs} & 312    & 1467   & 1779 & 97.29  & 92.74 &
 698    & 922   & 1620  & 98.18 & 83.33  & 
 1231   & 2370  & 3601  & 95.15 & 83.08\\ 
 LR-CNN \cite{Liu19_tnnls}  &1198    &680  &1878  &97.13 &92.58 & 
 1118   & 423   & 1541  & 98.27 & 85.36 & 
 1923 & 1460  & 3383 & 95.45 &84.83 \\ 
 Ms-CapsNet     &425    &779   &1204  &98.16 & 95.16 &
 468   & 407   & 875  & 99.02 &  91.22 & 
  1173  & 1798 & 2971  & 96.00 & 86.25\\

 \hline \hline
\end{tabular}
\label{table_result}
\end{table*}

In this subsection, the proposed method is compared with four closely related methods. The quantitative results of different methods on three datasets are displayed in Table \ref{table_result}. Fig. \ref{fig_result} exhibits the visual results of the proposed Ms-CapsNet together with all competitors.

Fig. \ref{fig_result}(d)-(h) present the change maps corresponding to the experiments reported in Table \ref{table_result}. On the Sulzberger dataset (the first row of Fig. \ref{fig_result}), the result of PCANet tends to be rather noisy, and it is afflicted with high FP value. Although other methods generally performed well, the results are deteriorated due to higher OE values. The proposed Ms-CapsNet exhibits less misclassified pixels and obtains the best PCC and KC values. 

On the Yellow River I and II datasets (the second and third rows of Fig. \ref{fig_result}), we can observe that the proposed Ms-CapsNet achieves at least 0.5\% improvement over other compared methods. Considering that the interference of different characteristics of speckle noise weakens the model performance, the proposed method is relatively noise robust. The PCANet suffers from high FP value, and there are many noisy regions in the generated change maps. LR-CNN performs better since local spatial restrictions can balance the influence of local noise. CNN-based methods can suppress noise interference to some extent through deep feature representation. However, relatively high OE values are still obtained. In general, the proposed Ms-CapsNet exhibits the best performance according to Table \ref{table_result} and Fig. \ref{fig_result}. It reveals that the proposed Ms-CapsNet benefits from the spatial relation exploration. 

\section{Conclusion}

In this paper, we proposed a new multiscale capsule network (Ms-CapsNet) for SAR image change detection. The proposed method benefits from two aspects: First, in order to enhance the spatial feature correlations, multiscale capsule module is utilized to model the spatial relationship of features between one object and another. Equivariant properties can be achieved by aggregating the feature from different positions. Further, we design an AFC module to alleviate the interference of speckle noise. The module can effectively convert pixel-wise intensities to activity local features. Extensive experiments are conducted on three real SAR datasets, and the experimental results demonstrated the superior performance of the proposed Ms-CapsNet.

\end{document}